\documentclass[runningheads]{llncs}

\usepackage{eccv}

\usepackage{eccvabbrv}

\usepackage{graphicx}
\usepackage{booktabs}

\usepackage[accsupp]{axessibility}  

\usepackage{hyperref}

\usepackage{orcidlink}

\usepackage{cite}
\usepackage{amsmath,amssymb,amsfonts}
\usepackage{algorithmic}
\usepackage{graphicx}
\usepackage{textcomp}
\usepackage{xcolor}
\usepackage{multirow}
\usepackage{glossaries}
\usepackage{textcomp}
\usepackage{svg}
\usepackage{booktabs}
\usepackage[inline]{enumitem}
\usepackage{blindtext}
\usepackage{siunitx}
\usepackage{overpic}
\usepackage{acro}
\usepackage{array}

\newacronym{cnn}{CNN}{Convolutional Neural Networks}
\newacronym{yolo}{YOLO}{You Only Look Once}
\newacronym{fps}{FPS}{frames per seconds}
\newacronym{sota}{SOTA}{state-of-the-art}
\newacronym{soc}{SoC}{System-on-Chip}
\newacronym{ble}{BLE}{\textit{Bluetooth} Low Energy}
\newacronym{cots}{COTS}{commercial of the shelf}
\newacronym{iot}{IoT}{Internet of Things}
\newacronym{aiot}{AIoT}{Artificial Intelligence of Things}
\newacronym{tpu}{TPU}{Tensor Processing Unit}
\newacronym{asic}{ASIC}{Application Specific Integrated Circuit}
\newacronym{fpga}{FPGA}{Field-Programmable Gate Array}
\newacronym{cpu}{CPU}{Central Processing Unit}
\newacronym{rbe}{RBE}{Reconfigurable Binaray Engine}
\newacronym{ip}{IP}{Intellectual Property}
\newacronym{qnn}{QNN}{Quantized Neural Networks}
\newacronym{dlau}{DLAU}{Deep Learning Accelerator Unit}
\newacronym{hd}{HD}{High Dimensional}
\newacronym{ar}{AR}{Augmented Reality}
\newacronym{ai}{AI}{Artificial Intelligence}
\newacronym{ml}{ML}{Machine Learning}
\newacronym{map}{mAP}{Mean-Average Precision}
\newacronym{pmic}{PMIC}{Power Management Integrate Ciruit}
\newacronym{ptq}{PTQ}{Post-Training Quantization}
\newacronym{slr}{SLR}{Single-Lens Reflex}
\newacronym{pcb}{PCB}{Printed Circuit Board}
\newacronym{nms}{NMS}{Non-maximum Suppression}
\newacronym{lora}{LoRa}{Long Range modulation technique}
\newacronym{fc}{FC}{Fabric Controller}
\newacronym{mcu}{MCU}{Microcontroller Unit}
\newacronym{sgd}{SGD}{Stochastic Gradient Descent}
\newacronym{mac}{MAC}{Multiply and Accumulate}
\newacronym{amp}{AMP}{Automatic Mixed Precision}
\newacronym{os}{OS}{Operating System}
\newacronym{csp}{CSP}{Cross-Stage Partial}

\definecolor{mydarkblue}{rgb}{0,0.08,1}

\usepackage{eso-pic}
\usepackage{url}
\AddToShipoutPictureBG*{
  \AtPageUpperLeft{%
    \put(0,-60){\raisebox{15pt}{\makebox[\paperwidth]{\begin{minipage}{21cm}\centering
      \textcolor{gray}{This paper has been accepted for publication at \\ ECCV 2024 Workshops, Milan, 2024.}
    \end{minipage}}}}%
  }
}

\begin{document}

\hyphenation{Tiny-issimo-YOLO}
\hyphenation{Tiny-issimo-YOLOv}


\title{Ultra-Efficient On-Device Object Detection on AI-Integrated Smart Glasses With TinyissimoYOLO} 

\titlerunning{On-Device Object Detection With TinyissimoYOLO}

\author{Julian Moosmann\inst{1}\orcidlink{0009-0007-0283-0031} \and
Pietro Bonazzi\inst{1}\orcidlink{0009-0006-6147-2214} \and
Yawei Li\inst{1}\orcidlink{0000-0002-8948-7892
} \and
Sizhen Bian\inst{1}\orcidlink{0000-0001-6760-5539} \and
\\ Philipp Mayer \inst{1}\orcidlink{0000-0002-4554-7937} \and
Luca Benini \inst{1,2}\orcidlink{0000-0001-8068-3806} \and
Michele Magno\inst{1}\orcidlink{0000-0003-0368-8923}}
\authorrunning{J.~Moosmann et al.}

\institute{ETH Zürich, 8092 Zürich, Switzerland \and
University of Bologna, 40126 Bologna, Italy}

\maketitle

\begin{abstract}
 Smart glasses are rapidly gaining advanced functions thanks to cutting-edge computing technologies, especially accelerated hardware architectures, and tiny \gls{ai} algorithms. However, integrating \gls{ai} into smart glasses featuring a small form factor and limited battery capacity remains challenging for a satisfactory user experience. 
To this end, this paper proposes the design of a smart glasses platform for always-on on-device object detection with an all-day battery lifetime. The proposed platform is based on GAP9, a novel multi-core \textit{RISC-V} processor from \textit{Greenwaves Technologies}. Additionally, a family of sub-million parameter TinyissimoYOLO networks are proposed. They are benchmarked on established datasets, capable of differentiating up to 80 classes on MS-COCO.
Evaluations on the smart glasses prototype demonstrate TinyissimoYOLO's inference latency of only 17ms and consuming 1.59mJ energy per inference. An end-to-end latency of 56ms is achieved which is equivalent to 18 \gls{fps} with a total power consumption of 62.9mW. This ensures continuous system runtime of up to 9.3 hours on a 154mAh battery. These results outperform MCUNet (TinyNAS+TinyEngine), which runs a simpler task (image classification) at just 7.3 \gls{fps}, while the 18 \gls{fps} achieved in this paper even include image-capturing, network inference, and detection post-processing.
The algorithm's code is released open with this paper and can be found here: \href{https://github.com/ETH-PBL/TinyissimoYOLO}{github.com/ETH-PBL/TinyissimoYOLO}
\end{abstract}

\glsreset{fps}
\glsreset{ai}
\glsreset{mcu}

\section{Introduction}
\label{sec:intro}

The rapid integration of advanced perception techniques into cutting-edge wearable computing devices has ushered in a transformative era, redefining how we engage with our surroundings and environment \cite{nahavandi2022application, bian2022exploring}. Among innovative wearables, smart glasses stand out as the next big thing in wearable computing \cite{sun2022design}. Their multifaceted applications span across a diverse spectrum, offering valuable support for professional applications \cite{spandonidis2023development}, while at the same time enhancing user experiences in entertainment and education\cite{iqbal2023adopting}, and most importantly, improving the quality of life for individuals with disabilities\cite{miah2018unique, ali2016smart, smart_glasses_for_visual_impaired}.

\begin{figure}[tb]
\centering
\begin{subfigure}{0.49\linewidth}
\includegraphics[trim={0 0cm 0 0cm},clip, width=\linewidth]{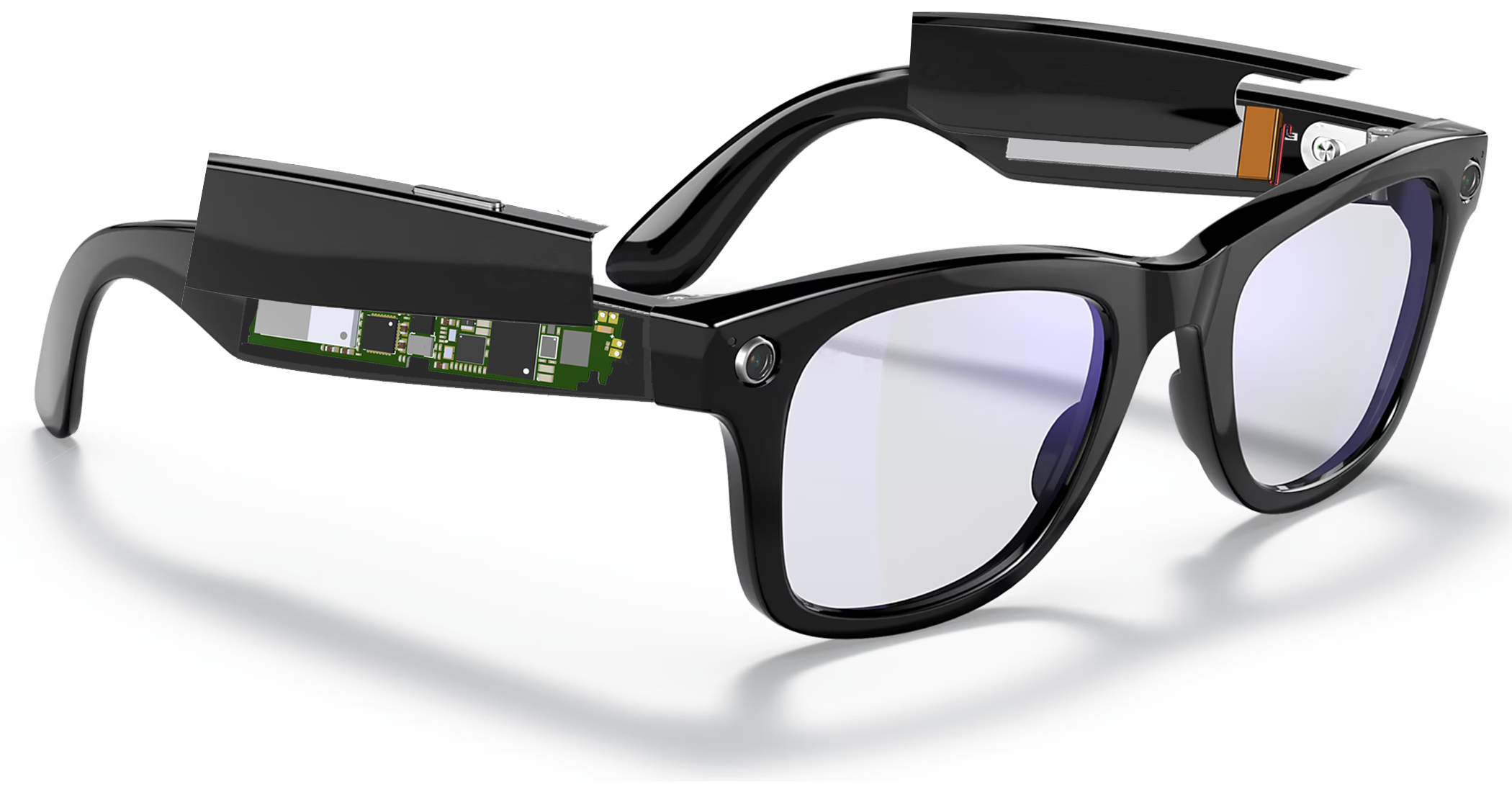}
\caption{\textbf{Smart Glasses with Electronics}:  The designed smart glasses hardware. The left temple holds the miniaturized electronics. The right temple contains the battery, with an energy content of up to \SI{154}{mAh}.}
\label{fig:smart_glasses_rendering2}
\end{subfigure}
\hfill
\begin{subfigure}{0.49\linewidth}
    \includegraphics[trim={0 0cm 0 0cm},clip, width=\linewidth]{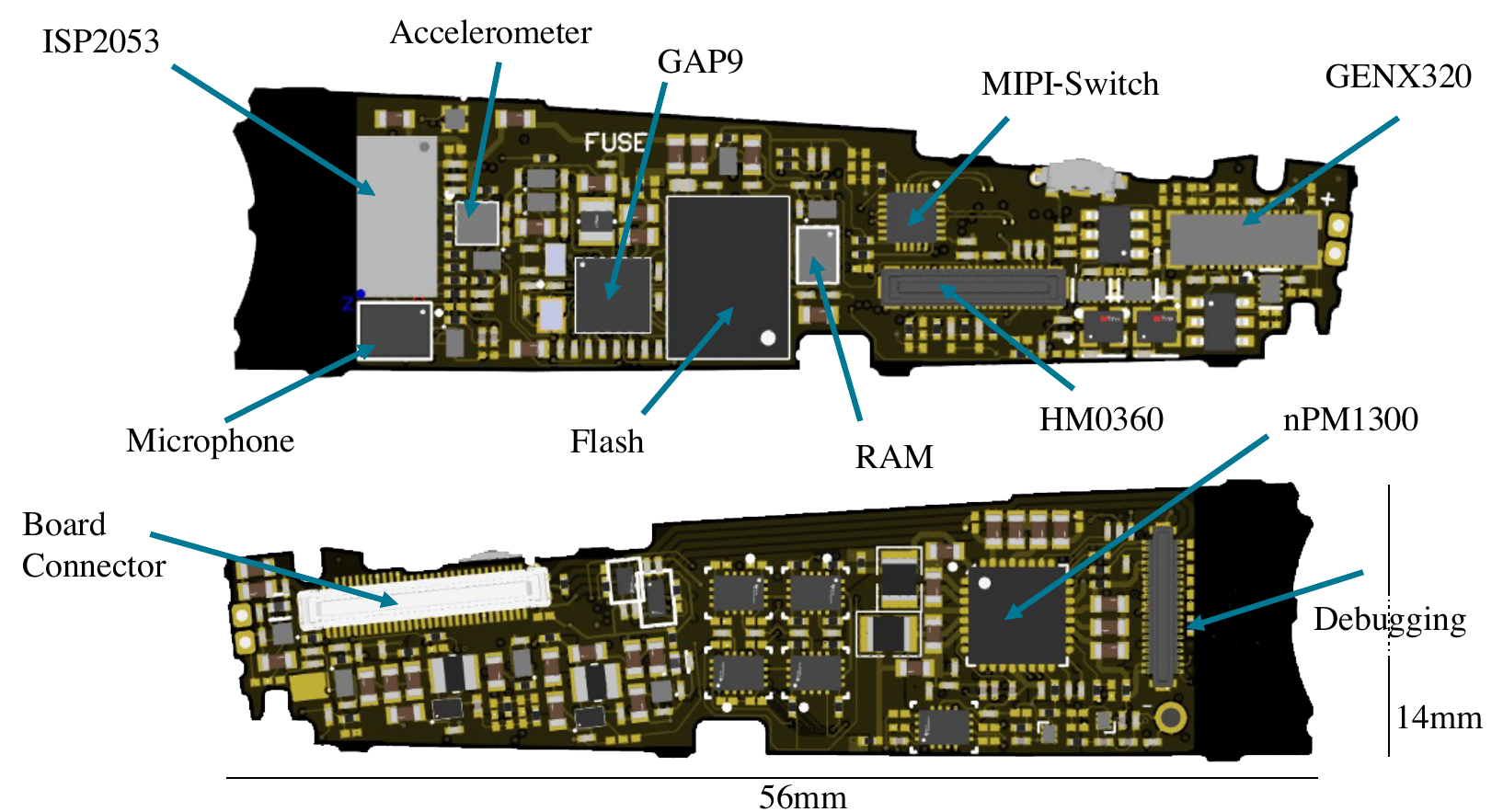}
\caption{\textbf{Miniaturized Smart Glasses Electronics:}  The proposed smart glasses platform based on GAP9 from \textit{Greenwaves Technologies.}}
\label{fig:smart-glasses-pcb}
\end{subfigure}
\caption{The designed smart glasses hardware, which retrofits commercial temples of smart glasses.}
\end{figure}
The mass adoption of current \gls{ai} technology is accentuating the trend in edge intelligence \cite{feng2019computer, wang2018embedded}, targetting computer vision \cite{feng2019computer, bonazzi2023tinytracker}, biomedical applications \cite{wang2022mi, moin2021wearable}, natural language processing \cite{zhao2023survey}, and many other spaces \cite{bian2022exploring}. 
Similar to environmental mapping \cite{mccormac20183dv, rosinol2020icra} and navigation \cite{crespo2020mdpi}, the detection and localization of objects \cite{huang2021bevdet, li2022bevformer} plays a pivotal role for understanding the visual context in which smart devices such as \gls{aiot} devices are operating.

Embedded machine learning for semantic scenery understanding \cite{smart_glasses_for_visual_impaired} in smart glasses enhances user interaction and brings intelligence right to the user's forehead while protecting user data, minimizing latency and energy. In particular, running object detection algorithms is a crucial functionality to enable smart decisions and information about the user's surroundings \cite{smart_glasses_for_visual_impaired, smart_glasses_object_detection}. 



Commercially available smart glasses such as the newly released \textit{RayBan-Meta}, \textit{Vuzix®} smart glasses family, are powered by \textit{Qualcomm}'s \textit{Snapdragon} AR1 and XR1 platforms with a power budget of a few hundred milliwatts. Despite the choice of using a relatively high-performance and powerful \gls{soc}, this capable processor is mostly used for transmitting the vast amount of data taken with high-resolution cameras and microphones. However, image or speech processing is only very partially done on-device\footnote{\url{https://about.fb.com/news/2023/09/introducing-ai-powered-assistants-characters-and-creative-tools/}}. This makes the 'smart' glasses primarily an input device for an \gls{ai} system working on the connected mobile phone or in the cloud \cite{smart_glasses_for_visual_impaired}.

To bridge the gap between the available computational resources and the demands of \gls{ai} algorithms, novel energy-efficient and ultra-low-power \glspl{mcu} with cluster cores for parallel processing, as well as dedicated accelerated hardware are being designed and are now available on the market\cite{conti2023marsellus, islamoglu2023ita, survey_extrem_edge_computing}. At the same time, new lightweight, and quantized networks are being proposed to achieve near \gls{sota} accuracy while having small-sized memory footprints\cite{moosmann2023flexible, bonazzi2023tinytracker} to fit the requirements of \glspl{mcu} and accelerators. Combining these novel multi-core \glspl{mcu} with a dedicated system design for ultra-low-power consumption enables a hardware platform for continuous \gls{ai} inference in smart glasses, with privacy protection and all-day long battery lifetime.

This paper presents the design and implementation of such an energy-efficient intelligent smart glasses system---see \cref{fig:smart_glasses_rendering2}---equipped with GAP9, an \gls{ai}-capable \gls{mcu} from \textit{Greenwaves Technologies}, which consumes power in the milliwatts range. We present the first end-to-end image processing pipeline on a \gls{mcu} which can process up to 18 \gls{fps} from image capturing to object localization on a \gls{mcu} device. Additionally, we propose an open family of quantized and lightweight TinyissimoYOLO \cite{moosmann2023tinyissimoyolo, moosmann2023flexible} networks with less than \qty{1}{MB} memory. All the contributions can be summarized as follows:

\begin{enumerate}
    \item  \textbf{Smart Glasses' System Architecture}: We propose a system architecture tailored for smart glasses applications with integrated ML acceleration. 
    
    \item  \textbf{Sub-Million Parameter YOLO Architectures}: We extend the TinyissimoYOLO series to support the latest datasets with up to 80 classes while having sub-million parameters when detecting more than 20 classes.

    \item  \textbf{End-to-end Real-Time Deployment}: We demonstrate the practical applicability of our proposed system by deploying the TinyissimoYOLO architectures on the smart glasses platform and by predicting images from the real world. 
    
    \item  \textbf{Power Efficiency and Performance Evaluation}: We conduct extensive experiments to validate the power efficiency and performance of our smart glasses system. Comparative analyses against commercial edge vision systems reveal superior energy efficiency and longer battery life for our system.

    \item  \textbf{Open-Source Implementation}: To facilitate reproducibility and encourage further research, we release the source code of our optimized TinyissimoYOLO architecture versions, allowing the community to build upon our work and extend it to new application domains.

\end{enumerate}

The paper is structured as follows: in \cref{sec:related_work}, we investigate recent related works regarding both the \gls{sota} object detection neural networks and their deployability onto \glspl{mcu} using different frameworks. Then, the extension of the new TinyissimoYOLO versions are described in \cref{sec:neural_network}. \cref{sec:system_design} introduces the prototype design of the smart glasses aiming for latency and energy efficiency during onboard AI execution.  The results are presented in \cref{sec:results}, including the detection results and the system evaluation of GAP9 integrated with the proposed smart glasses system.
Finally, we conclude our work in \cref{sec:conclusion}.


\section{Related Work} 
\label{sec:related_work}

The following section provides an overview of \gls{sota} object detection algorithms more specifically for use with edge processors. Additionally, we discuss different deployment frameworks to deploy networks on edge devices. Lastly, the currently available smart glasses on the market, as well as research projects, are summarized and set into perspective while this work is further motivated.
 
\subsection{SOTA Object Detection on Microcontrollers}
\gls{yolo} \cite{redmon_you_2016}, is an optimized deep learning algorithm used to perform real-time object detection on GPU-class devices \cite{jiang2022review}. It utilizes a feature extraction \gls{cnn} backbone and detection head to perform localization of the extracted semantic information. Meanwhile, there exist a family of different \gls{yolo} versions which differ in backbone, head structure, network size, inter-network connection, and used layer operations \cite{redmon2017yolo9000, redmon2018yolov3, bochkovskiy2020yolov4, li2022yolov6, wang2023yolov7, Jocher_YOLO_by_Ultralytics_2023, Jocher_YOLOv5_by_Ultralytics_2020, wang2024yolov10}. However, even the smallest \textit{"nano"} versions of the YOLO-family exhibit roughly 3 million parameters, making them unsuitable for \glspl{mcu}. As such, TinyissimoYOLO \cite{moosmann2023tinyissimoyolo} and its successor \cite{moosmann2023flexible} try to bridge the gap between accuracy and network size \cite{saha2022sensors} while maximizing the available compute acceleration on the milliwatt edge device. Until now, TinyissimoYOLO was not able to detect 20 or more classes while fitting seamlessly on a \gls{mcu}. The earlier version's output layer scaled linearly with the number of detection classes, quickly reaching 2 million parameters and more. Therefore, this work increases the number of detection classes while maintaining the network size below \qty{1}{MB}.

Networks such as YOLOX-Nano \cite{ge2021yolox}, PP-PicoDet \cite{yu2021pp} and NanoDet-M\footnote{https://github.com/RangiLyu/nanodet} achieve higher mean-average precision while utilizing specialized network layers to decrease network parameters. While PP-PicoDet and NanoDet-M incorporate depth-wise convolutional layers and enhanced ShuffleNet \cite{zhang2018shufflenet} blocks, they fail to fully exploit the hardware acceleration built into \glspl{mcu} such as GAP9. Conversely, YOLOX-Nano encounters memory constraints of \glspl{mcu} due to its large input resolution of 640 pixels, which RGB image costs more than \qty{1}{MB} of memory solely for the input image. Consequently, simple convolutional layers with a kernel of 3x3 are preferable for current accelerators built into \gls{mcu} class devices\cite{conti2023marsellus}, which makes TinyissimoYOLO favorable for on-device execution. Nonetheless, deploying a battery-powered device with a camera resolution under 300 pixels to distinguish 80 classes of MS-COCO \cite{lin2014microsoft} is impractical. Therefore, we focus our evaluation of the networks on PascalVOC \cite{everingham2010pascal} and evaluate a few on MS-COCO for a fair comparison against similar-sized networks. Lastly, we compare them to MCUNet, which---similar to us---reports an end-to-end system latency deployed on \gls{mcu} devices. For a comprehensive comparison, refer to \cref{tab:yolo_comparison}. To the best of the author's knowledge, the only work that implemented an end-to-end object detection pipeline on a microcontroller is MCUNet \cite{lin_mcunet_2020}. They claim to have 10 \gls{fps} inference execution. When considering the additional time needed for capturing an image, MCUNet's end-to-end latency achieves 7.3 \gls{fps}.

\begin{table}[!b]
    \caption{\textbf{Network Comparison:} Overview of similar-sized networks compared to our TinyissimoYOLO versions and evaluated on PascalVOC and MS-COCO.}
    \centering
    \label{tab:yolo_comparison}
    {\scriptsize
    \begin{tabular}{@{} 
     >{\raggedright\arraybackslash}p{0.25\linewidth} 
     >{\raggedleft\arraybackslash}p{0.1875\linewidth} 
     >{\raggedleft\arraybackslash}p{0.1875\linewidth} 
     >{\raggedleft\arraybackslash}p{0.1875\linewidth} 
     >{\raggedleft\arraybackslash}p{0.1875\linewidth} 
  @{}}\toprule
        \textbf{Model} & \textbf{Image}& \textbf{Parameters} & \textbf{PascalVOC} & \textbf{MS-COCO} \\
        & \textbf{Resolution}& \textbf{(M)} & \textbf{mAP@(50-95)} & \textbf{mAP@(50-95)} \\
        \midrule
        MbV2+CMSIS\cite{lin_mcunet_2020} & 128 & 0.87 & 32\% & \\
        MCUNet\cite{lin_mcunet_2020} & 224 & 1.2 & 51\% & \\
        MCUNetV2-M4\cite{lin_mcunetv2_2021} & 224 & 1.01 & 65\% & \\
        MCUNetV2-H7\cite{lin_mcunetv2_2021} & 224 & 2.03 & 68\% & \\
        NanoDet-M$^1$ & 320 & 0.95 &  & 21\% \\
        YOLOX-Nano \cite{ge2021yolox}& 640 & 0.91 & & 26\% \\ 
        PP-PicoDet \cite{yu2021pp} & 320 & 0.99 & & 27\% \\ 
        \midrule
        TY-v1 3cls\cite{moosmann2023tinyissimoyolo} & 224 & 1.66 & 68\% & \\ 
        TY-v1 10cls\cite{moosmann2023flexible}  & 224 & 2.36 & 65\% & \\ 
        TY-v1 20cls\cite{moosmann2023flexible}  & 224 & 3.35 & 60\% & \\ 
        \midrule
        TY-v1.3-Small$^*$ & 256 & 0.40 & 30\% & \\
        TY-v1.3-Big$^*$ & 256 & 0.96 & 38\% & \\
        \midrule
        YOLO-v5-nano\cite{Jocher_YOLOv5_by_Ultralytics_2020} & 640 & 2.66 & & 34\%\\
        TY-v5-Small$^*$ & 256 & 0.63 & 35\% & \\
        TY-v5-Big$^*$ & 256 & 0.89 & 42\% & 14\% \\
        \midrule
        YOLO-v8-nano\cite{Jocher_YOLO_by_Ultralytics_2023} & 640 & 3.2 & & 37\% \\
        TY-v8-Small$^*$ & 256 & 0.71 & 39\% & \\
        TY-v8-Big$^*$ & 256 & 0.84 & 44\% & 15\% \\
        \midrule
        TY-v10$^*$ & 256 & 0.85 & 49\% & 14\% \\ 
        \bottomrule
        \multicolumn{5}{l}{$^*$ This works' TinyissimoYOLO networks deployed and evaluated on GAP9.} \\
        \multicolumn{5}{l}{$^1$ \url{https://github.com/RangiLyu/nanodet}}
    \end{tabular}
    }
\end{table}

To achieve lower end-to-end latency this work builds on the results published in the TinyissimoYOLO papers\cite{moosmann2023tinyissimoyolo, moosmann2023flexible}. They compared several ARM-Cortex M4 and M7 \glspl{mcu} from STMicroelectronics,  Apollo4b from \textit{Ambiq} (an ultra-low-power MCU using sub-threshold technology), and the MAX78000 \gls{mcu} from \textit{Analog Devices}. The work exploits a \gls{sota} hardware accelerator for \gls{cnn} networks, and a \textit{RISC-V} low-power parallel processor with hardware accelerator, GAP9 from \textit{Greenwaves Technologies}. It showed that MAX78000 is parallelizing the compute workload best. Nevertheless, GAP9 achieves the same energy efficiency despite having two-fold less MAC/cycle. Additionally, the papers show that the increased clocking frequency of GAP9 together with its flexible combination of cluster cores and neural engine, allows for larger, less restricted networks while being able to parallelize the overall workload better.

Therefore, this paper leverages the GAP9's exceptional energy-efficient parallel processing capabilities and integrates it with a low-power \gls{ble} transceiver chip. This combination forms the backbone of our design to facilitate robust object detection in smart glasses with seamless connectivity.

\subsection{Smart Glasses}
Smart glasses---or in-general \gls{ar}---focus on a general computing concept \cite{liu2022augmented} to process the user-device interaction while communicating with cloud or smartphones via Bluetooth and Wi-Fi \cite{8368051}. Big tech companies such as \textit{Google}\footnote{\url{https://developers.google.com/glass-enterprise/}} or \textit{Apple}\footnote{\url{https://developer.apple.com/visionos/}}, further rely on community APPs being developed and processed on the device, requiring the manufacturer to provide an easy-to-develop software stack, to abstract the hardware from the software. This requires an \gls{os} environment to run at reasonably fast speeds, such that user interactions are ensured to run smoothly. In particular, these requirements hinder smart on-device sensor data processing, resulting in the data being processed in the cloud, while data and user privacy are not absolutely contained and guaranteed. 

In contrast, several research smart glasses projects aim at the counterpart by running \gls{ai} algorithms on the device \cite{sun2022design, hong20152}. Others investigate the human-machine interaction with smart glasses \cite{salvaro2018wearable}. However, many publications focus on a single smart glasses application scenario and design the system accordingly, e.g., for visually impaired people \cite{miah2018unique, ali2016smart} or smart gadget aid for medical \cite{8962044, subhan2023ai} or construction work \cite{danielsson2020augmented}.
Nonetheless, to the best of our knowledge, none of the proposed research prototypes are integrated into the thin frames of the passive glasses \cite{scherer2022widevision} and will therefore not be non-stigmatizing  nor fashionable, in contrast to actual smart glasses products such as the \textit{RayBan-Stories}\footnote{\url{https://tech.facebook.com/2023/2/the-making-of-ray-ban-stories/}} 
or the brand-new \textit{RayBan-Meta}\footnote{\url{https://about.fb.com/news/2023/09/new-ray-ban-meta-smart-glasses/}}.
Therefore, this work introduces electronics that retrofit a functional and fashionable smart glasses frame with a peak power consumption below \qty{100}{mW}. This makes our solution to smart glasses not only aesthetically appealing but also capable of directly executing demanding \gls{yolo} object detection tasks, showcasing both efficiency and effectiveness in image processing for real-world applications. 


\section{TinyissimoYOLOs} 
\label{sec:neural_network}

\label{sec:networks}
This section presents a family of sub-million parameter detection algorithms, based on a different version of the \gls{yolo} architecture, which we have developed for accelerated \glspl{mcu} or in general low-power edge processors. These networks are proposed as a trade-off between computational resources and performance, as for example in on-device execution for smart glasses systems. The networks predict multi-object-class probabilities and bounding boxes from 256x256 resolution images. However, the networks can be adapted for larger and smaller resolutions. 

\subsection{Network Architectures} 

\label{chap:tinyissimoyolo}
YOLOv5, YOLOv8, and YOLOv10 employ different backbone and head architectures to predict class probabilities and bounding boxes. We accurately evaluated the respective versions to extract their performance under sub-million parameter constraints, see the network specifications in \cref{tab:yolo_comparison}. This has been conducted to establish a family of networks suitable to be deployed on \gls{aiot} devices. In particular, for the deployment of multi-object detection networks on a \gls{mcu} with ML acceleration, such as the GAP9. All the networks described below have been deployed on the GAP9 and not only will their detection capability be evaluated, but also their deployed inference performance, energy consumption, and their ability to parallelize the inference execution on such a hardware.

The new TinyissimoYOLOv1.3 key differentiator lies in the incorporation of the Detection Block from the V8 architecture into the final prediction layer of TinyissimoYOLOv1.3. In contrast to the originally proposed TinyissimoYOLO versions \cite{moosmann2023flexible, moosmann2023tinyissimoyolo} this strategic enhancement imbues the latter with heightened sensitivity and a larger number of detection parameters, without scaling the detection block linearly. Consequently, TinyissimoYOLOv1.3 demonstrates precise multi-class object detection capability on sub-million parameters.

TinyissimoYOLOv5's architecture features a \gls{csp} connection \cite{cspnet} bottleneck module named 'C3'. In C3, the input is duplicated through two separate 1x1 convolutions which are then concatenated and processed through a final 1x1 convolution to produce the final output. The main distinction between the architecture of YOLOv5 and YOLOv8 lies in the number of convolutions and on the expansion or contraction of the hidden channels in the \gls{csp}. Differently from YOLOv5, TinyissimoYOLOv5-big has a layer channel multiple of 0.15 instead of 0.25 (YoloV5-nano) and contains 4 times fewer parameters, while the small version uses a channel multiple of 0.1.

YOLOv8 \gls{csp} Block called 'C2F', which powers TinyissimoYOLOv8 starts with a 1x1 convolution that expands the input channels to twice the hidden channel size. Then, it splits the output into two equal parts and applies a series of bottleneck layers on each part. Finally, it concatenates the processed outputs and applies a second 1x1 convolution. TinyissimoYOLOv8 has a depth multiple of 0.30 instead of the 0.33 in YoloV8-nano, a layer channel multiple of 0.18 instead of 0.25, and contains 5 times fewer parameters than the nano version, while the small version uses a channel multiple of 0.1.

The latest \gls{yolo} version v10 \cite{wang2024yolov10}, removes the need to perform \gls{nms} during training and improves the detection accuracy on established datasets. The enhanced \gls{csp} feature extraction backbone, combined with the updated and \gls{nms}-free head achieves \gls{sota} detection performance while having faster execution times.
Our TinyissimoYOLOv10, utilizes the same backbone, neck, and head however, we use a channel multiple of 0.18 and a depth multiple of 0.15 to achieve sub-million number of parameters.

\subsection{Implementation Details}

We trained the proposed models on an \textit{NVIDIA} \textit{GeForce RTX} 4090 for $1000$ epochs, with a batch size of 64. The initial learning rate $lr=1e-3$ was reduced using a cosine learning rate scheduler to $lr=1e-5$ after a 3 epoch-long warmup phase $(lr=1e-2)$. The training process utilised a Multi-Object Detection Loss \cite{redmon_you_2016}, and images with a resolution of 256$\times$256 pixels. Additionally, several established image augmentations are applied to the images, such as exposure and saturation adjustments in the HSV color space, horizontal image flipping, image translation, scaling as well as image mosaicing. We optimize the weights using \gls{sgd} $(\text{momentum}=0.937)$ and trained with \gls{amp}. The networks have been integrated into the \textit{Ultralytics} framework \cite{Jocher_YOLO_by_Ultralytics_2023}. Thereafter, we applied \gls{ptq} \cite{han_deep_2016} to quantize the networks to 8-bit integer values, such that the sub-million parameter network fits into a \gls{mcu} memory of \qty{1}{MB}.


\section{System Design} 
\label{sec:system_design}

\begin{figure*}[tb]
\begin{center}
 \begin{overpic}[width=0.95\linewidth]{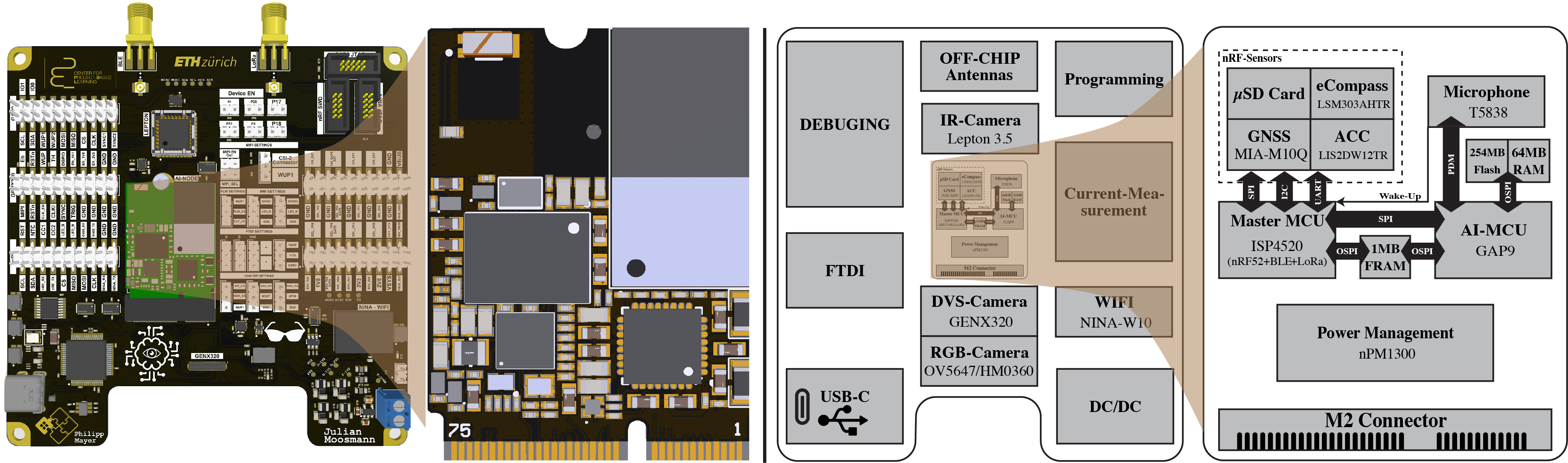}
        \put(25.5,-2){\textcolor{black}{(a)}}
        \put(74.5,-2){\textcolor{black}{(b)}}
    \end{overpic}
\caption{\textbf{Development Board:}  a) The proposed hardware system consists of two boards. The board on the left---development board---is shown, featuring additional power circuitry, multiple camera interfaces, Wi-Fi, and debug possibility. The smart glasses board---zoomed in---, features two \glspl{mcu}, several sensors, and a power management system for stand-alone operation. b) shows the hardware block diagram for the development and smart glasses board respectively.}
\label{fig:hw_overview}
\end{center}
\end{figure*}

To achieve \textit{"truly"} smart and energy-efficient smart glasses, we propose a hardware-software smart glasses solution, eventually eliminating the transmission of private data, and decreasing inference latency while effectively increasing battery run-time. For this, a development board was designed---see \cref{fig:hw_overview}a---and utilized for rapid end-to-end evaluations while the new smart glasses PCB has been designed to retrofit existing smart glasses temples for the final system, see \cref{fig:smart_glasses_rendering2}. 

\subsection{Smart Glasses Design}
 A modular design was targeted so that reuse of the same platform can be guaranteed for other ultra-low-power all-day battery runtime \gls{aiot} applications. All components are tightly integrated onto a custom,  miniaturized \gls{pcb}, to replace existing \gls{pcb} of commercial smart glasses.
 
The design features dedicated low-power infrastructure to power cycle energy-hungry sensors while powering the minimal required components. This platform is co-designed with a development platform for the rapid integration of further research sensors.

\cref{fig:hw_overview}a shows the hardware overview. 
The development board is shown on the left, while the smart glasses electronics is the zoomed-in \gls{pcb} (green) on the development board. The right side of the \cref{fig:hw_overview}b describes the hardware using a block diagram. The left side shows the development board, while the zoomed-in version is the smart glasses electronics. The M2 connector aims for development and reusability. To further minimize the form factor and fit temples of smart glasses, several on-board sensors are left away, resulting in the design that can be seen in \cref{fig:smart-glasses-pcb}. The final miniaturized smart glasses \gls{pcb} fits smoothly inside the temples of smart glasses, see \cref{fig:smart_glasses_rendering2}.

\paragraph{Smart Glasses Electronics: } The platform can be broken down into the following three main parts: 
\begin{enumerate*}[label=(\roman*),,font=\itshape]
    \item battery and power management,
    \item dual-\gls{mcu} architecture leveraging energy-efficiency
    \item various sensors such as accelerometer, multiple cameras, and microphones.
\end{enumerate*} 

The battery and power management is needed for untethered operation of the smart glasses' electronics. One ultra-low-power \gls{mcu} is in charge of wireless communication, low-power signal processing (i.e. data from MEMS sensors), and power management of the various power domains by using multiple power switches in series and parallel within the same voltage levels. 
The aforementioned tasks are controlled by a \gls{soc}. In the development board, the ISM4520 has been chosen. It features a low-power nRF52 from \textit{Nordic Semiconductor} with a built-in \textit{ARM} Cortex-M4 processor, on-chip \gls{ble} and \gls{lora}. The smart glasses, however, don't need \gls{lora}. Therefore the ISP2053 has been chosen. It features a dual-core nRF53 with an embedded \textit{Bluetooth} 5.2 module. This \gls{soc}'s footprint is smaller than the one built into the devboard, making it ideal for the smart glasses system.

In addition to the \gls{soc} with built-in communication modules, a more power and energy-efficient \textit{RISC-V} parallel processor, GAP9 from \textit{Greenwaves Technologies}, is used for enabling computational intensive on-the-edge \gls{ai} algorithm inference for image and audio processing using neural networks. 
GAP9 enables both parallelization and hardware acceleration while the GAP9 achieves the overall best trade-off performance in a few milliwatts envelope, considering latency, energy consumption per inference, and parallelization of the computation. 
In particular, the GAP9 \gls{soc} has a built-in general-purpose neural network accelerator capable of running \gls{sota} algorithms' operations whenever the algorithm fits inside the memory. A FRAM is placed such that both GAP9 and nRF52 can access a shared memory space to share sensor data and system states. Additionally, GAP9 has a big amount of off-chip memory such as \SI{254}{MB} of Flash and \SI{64}{MB} DRAM. 

The miniaturized smart glasses electronics shown in \cref{fig:smart-glasses-pcb}, is a minimal electronics design, incorporating 2 cameras---GENX320 from Prophesee and HM0360 from Himax---, an accelerometer---LIS2DW---, a microphone and the above-described power management as well as \gls{ble} 5.2. The miniaturized design requires a high-precision \gls{pcb} design with 8-impedance matched layers for the camera's CSI-2 protocol connection and the OctoSPI interface between GAP9 and external RAM and FLASH.  
However, for system development, the M2 \gls{pcb} was utilized which features additional sensors, such as the eCompass---LSM303---, GNSS module---MIA-M10Q---, Microphones with wake-up capabilities---T5838---and an on-board micro-SD card storage for gigabytes of data collection. Removing these additional sensors of the M2 \gls{pcb} allows all the electronics design to be shrunk into a trapezoidal form factor, \cref{fig:smart-glasses-pcb} of \SI{14}{mm} x \SI{56}mm with a height of \SI{3}{mm} to fit the temples, see \cref{fig:smart_glasses_rendering2}.

\textit{The development board}---see \cref{fig:hw_overview}---is designed for modular, rapid design and integration of various new sensors and system development. The main focus is on the design of GAP9's interfacing with different cameras using the CSI-2 protocol. The single-line CSI-2 interface is multiplexed to interface two cameras. Once a connector for interfacing \gls{cots} RGB cameras, designed for use with Raspberry Pis has been added. This interface is shared with a second camera connector, specifically designed for the new event-based camera GENX320 from Prophesee. As a third camera option, a Lepton3.5 is added to the development board for adding infrared camera capabilities and interfaced via SPI.

\subsection{End-to-end System and Experimental Settings}
First, we describe which hardware was used for system deployment, followed by the deployment procedure of the developed TinyissimoYOLO networks. Lastly, we describe the end-to-end system and how the experiments are conducted.

To showcase onboard intelligence and longer battery runtime, an end-to-end firmware has been designed and deployed on the GAP9 hardware. For the power evaluation, the development board together with the M2 \gls{pcb} is used. Since the M2 system features the same hardware used for the end-to-end system, power measurements have been conducted with it. To maximize similarity, the sensors and peripherals not fitted in the miniaturized smart glasses \gls{pcb} are switched off.

The deployment of the networks is done using \textit{Greenwaves Technologies}' NNTool. Starting with the unquantized exported .onnx file from the \textit{Ultralytics} framework, NNTool has an integrated \gls{ptq} flow. As such, the networks were deployed using 8-bit integer precision. Once quantized, the network can be auto-tiled by \textit{Greenwaves Technologies}' Autotiler. This tool is used for automatic code generation, in particular, to generate C-code for the deployment onto the target cores or accelerator (NE16) of GAP9. Further, the user can specify the amount of L1, L2, and L3 memory to be used and the Autotiler tool will generate the network according to the constraints given such that the network is executed with minimal memory wait stalls, depending on each layer's size and operation used. Once the C-code has been generated, the network can be integrated into a custom user-specific project.

\begin{figure}[tb] 
\centering
    \includegraphics[trim={0 0cm 0 0cm},clip, width=0.85\linewidth]{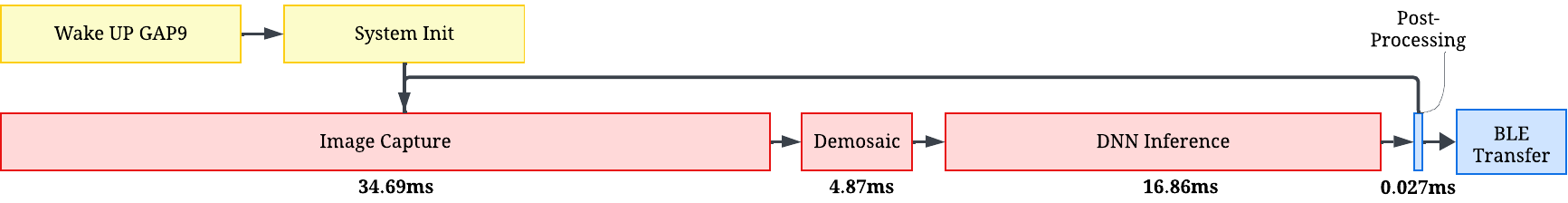}
\caption{\textbf{End-to-End System Overview:}  The image shows the flow chart of the demonstrator firmware, including the execution latency for the corresponding task. The box sizes are in relative size to the execution time.} 
\label{fig:demo-overview} 
\end{figure}

We implemented the end-to-end pipeline which is visualized in \cref{fig:demo-overview}. First, a raw image in a Bayern pattern is captured. The image gets demosaiced on the \gls{fc} of GAP9. TinyissimoYOLO is used to predict objects while its network output is post-processed to extract the bounding boxes of the detected objects. The image-capturing process uses double-buffering, to decrease latency arising from the image sensor data acquisition. The demosaic process as well as the post-processing of the network's output utilizes the \gls{fc}, while the object detection network runs on the cluster and on the neural engine.

The measurements of the networks are done by measuring the current consumption over 10 consecutive network inferences and calculating the average current consumption. Knowing the voltage level at which GAP9 is operating, the power as well as the energy consumption is calculated. We also calculate the number of \gls{mac} the network has to conduct one inference. 

Lastly, the full end-to-end system is measured on board level, to incorporate the camera, both microcontrollers, the \gls{pmic}, and all the voltage converter's inefficiencies.

\section{Results} 
\label{sec:results}
This chapter summarizes the results achieved in this work. First, the networks' detection accuracy is presented. Second the networks are deployed on the GAP9 and the achieved performance is reported. Lastly, the full end-to-end system results are provided including an always-on battery runtime estimation for the end-to-end system. 

\subsection{Detection Results}

In this section, the performance evaluation of the various TinyissimoYOLO network configurations on the PascalVOC test dataset is presented. The results are summarized in \cref{tab:yolo_comparison} and visualized in \cref{tab:yolo_prediction} using images taken with GAP9---first row of \cref{tab:yolo_prediction}---, and an \gls{slr} camera---second row. The evaluation of the newly proposed TinyissimoYOLO network with the different network configurations on the PascalVOC dataset reveals a noteworthy improvement in \gls{map} scores. Specifically, transitioning from TinyissimoYOLOv1.3-Small to TinyissimoYOLOv8-Big or the TinyissimoYOLOv10 architecture results in a substantial \gls{map} enhancement, showcasing the effectiveness of the latter in accurately detecting objects across various classes. 

\begin{table*}
    \centering
    \caption{\textbf{Qualitative Results:} TinyissimoYOLOv8 running on images captured with our system (row 1) and with images taken using an SLR camera (row 2).}
    \resizebox{0.85\linewidth}{!}{%
    \begin{tabular}{cccccc}
    \rotatebox[origin=c]{90}{\textbf{GAP9}} &
        \raisebox{-0.5\height}{\includegraphics[width=3cm, height=3cm]{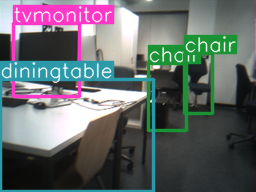}} & 
        \raisebox{-0.5\height}{\includegraphics[width=3cm, height=3cm]{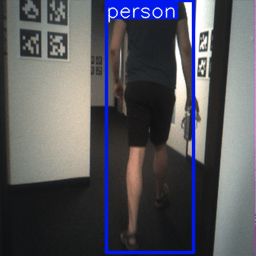} }& 
        \raisebox{-0.5\height}{\includegraphics[width=3cm, height=3cm]{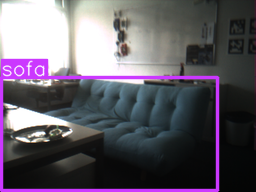} }& 
         \raisebox{-0.5\height}{\includegraphics[width=3cm, height=3cm]{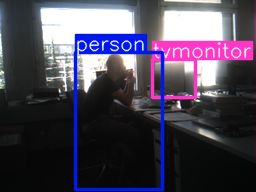}} &
       \raisebox{-0.5\height}{ \includegraphics[width=3cm, height=3cm]{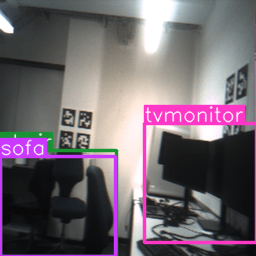} }\\
        \\ [0.1cm]
        \rotatebox[origin=c]{90}{\textbf{RLC Camera}} &  
        \raisebox{-0.5\height}{\includegraphics[width=3cm, height=3cm]{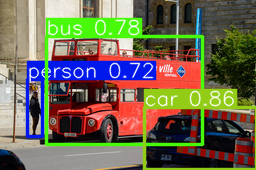}} & 
        \raisebox{-0.5\height}{\includegraphics[width=3cm, height=3cm]{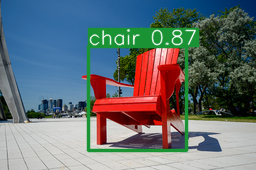}}& 
        \raisebox{-0.5\height}{\includegraphics[width=3cm, height=3cm]{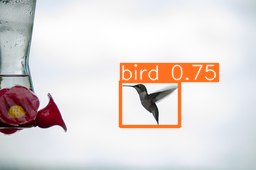}} & 
         \raisebox{-0.5\height}{\includegraphics[width=3cm, height=3cm]{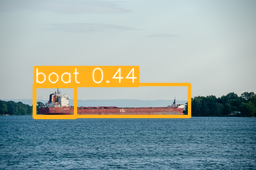}} &
        \raisebox{-0.5\height}{\includegraphics[width=3cm, height=3cm]{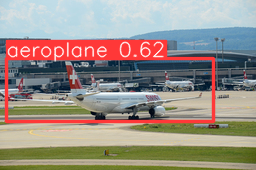}}\\
        \\ [0.1cm]
    \end{tabular}
    }
    \label{tab:yolo_prediction}
\end{table*}

\subsection{Deployed Networks on GAP9}
\label{sec:network_performance}
\begin{figure}[tb] 
\centering
\begin{subfigure}{0.4\linewidth}
\includegraphics[trim={0.5cm 0cm 0.5cm 0cm},clip, width=\linewidth]{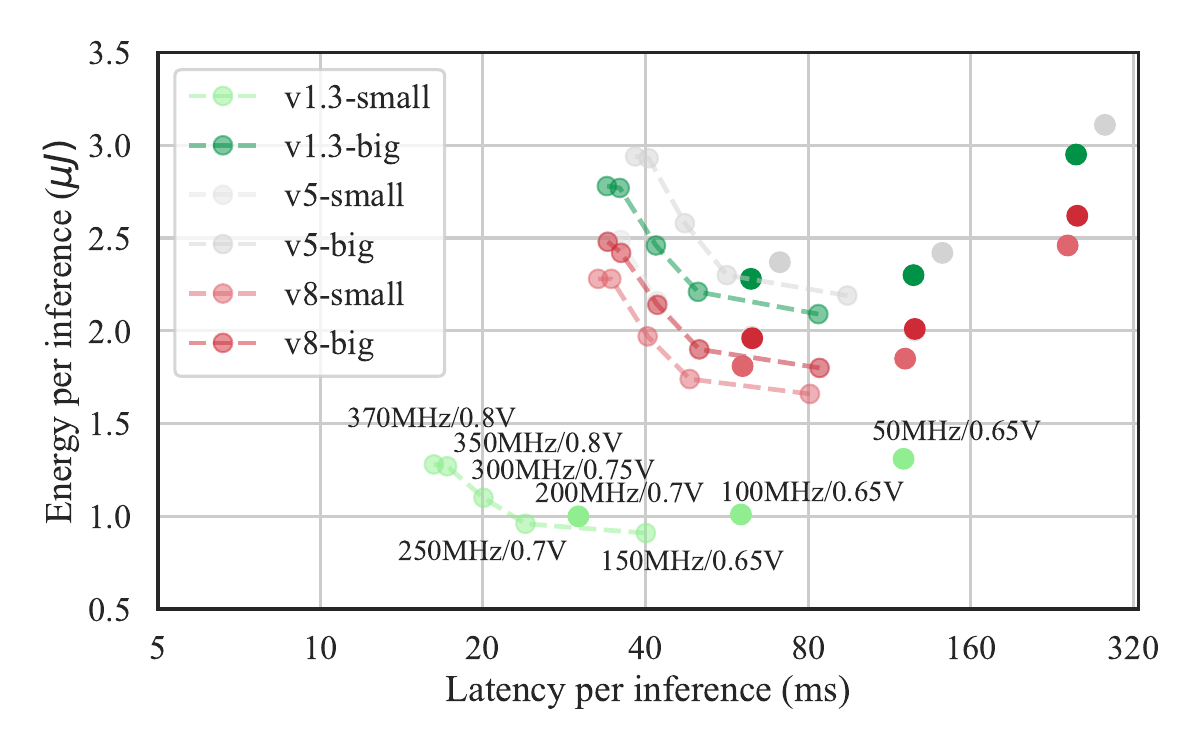}
\caption{\textbf{Latency vs. Energy:}  TinyissimoYOLOv1.3, v5 and v8 quantized deployed on GAP9's NE16 running at different voltage levels and different clocking frequencies. }
\label{fig:sweep} 
\hfill
\end{subfigure}
\begin{subfigure}{0.59\linewidth}
\centering
\includegraphics[trim={0.25cm 0.25cm 0.25cm 0.25cm},clip, width=0.65\linewidth]{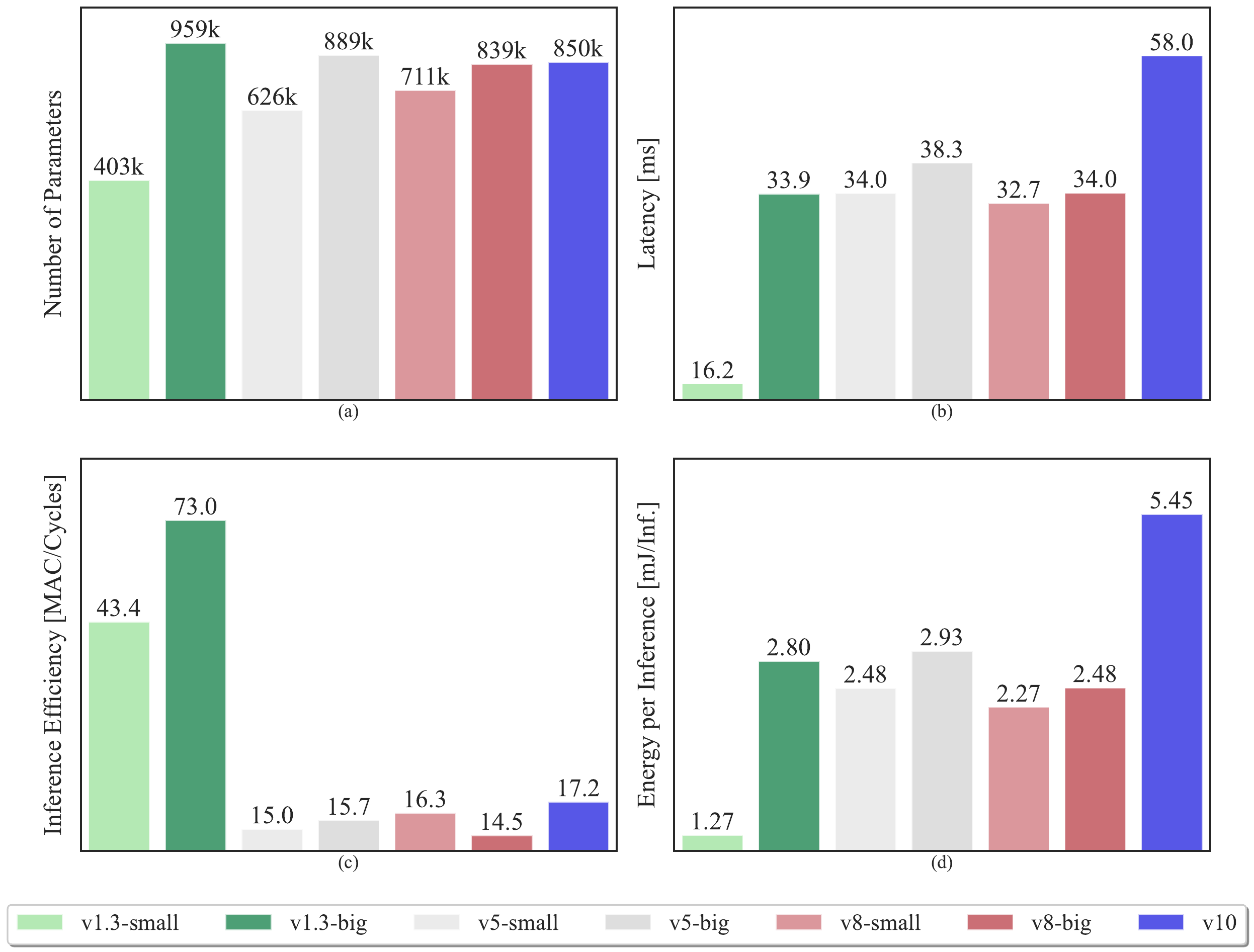}
\caption{\textbf{Deployed TinyissimoYOLO Versions (@0.8V; 370MHz):} This comparison shows a) the number of parameters of each TinyissimoYOLO version b) the latency of each network, c) the inference efficiency, or energy consumption per inference and d) the MAC/cycle, or parallelization workload. }
\label{fig:tinyissimoyolos_on_gap9}
\end{subfigure}
\caption{Evaluation of the Deployed TinyissimoYOLO Versions on GAP9}
\end{figure}

The networks are 8-bit quantized and deployed on the NE16 accelerator of GAP9 and compared in terms of latency---\cref{fig:tinyissimoyolos_on_gap9}-b---, inference efficiency---\cref{fig:tinyissimoyolos_on_gap9}-c---, and energy per inference---\cref{fig:tinyissimoyolos_on_gap9}-d---for each TinyissimoYOLO network version v1.3, v5, v8-small/big and v10. TinyissimoYOLOv1.3 outperforms the other networks in terms of latency with only \SI{16.9}{ms} execution time. V5 and both v8 variants need \SI{32.7}{ms}, \SI{34}{ms}, and \SI{36.6}{ms}, respectively. In terms of inference efficiency, v1.3 is best parallelizable with up to \SI{43.37}{MAC/cycle}. The v5 and v8 versions have \SI{16.28}{MAC/cycle}, \SI{14.98}{MAC/cycle}, \SI{15.27}{MAC/cycle}, respectively. 
Since v1.3 performs fastest and most parallelized, it's also the most energy efficient consuming only \SI{1.27}{mJ}, followed by v5 consuming \SI{2.34}{mJ}, v8-small with \SI{2.48}{mJ} and v8-big consuming \SI{2.62}{mJ}.

 Deploying the networks at different clocking frequencies of the NE16 core at different voltage levels leads to a Pareto Optima curve for running the networks quantized on the \gls{soc} of GAP9. \cref{fig:sweep} shows the TinyissimoYOLOv1.3, v5, and v8 being deployed in both 'small' and 'big' versions. It shows that, while v1.3-Small has the lowest detection robustness, as seen in \cref{tab:yolo_comparison}, it benefits from the lowest inference latency and has the lowest energy consumption. Additionally, the frequency sweep for all the TinyissimoYOLO versions proposed can be used as a look-up table for finding the ideal network's execution frequency. Depending if the application allows for low energy consumption and low accuracy, running TinyissimoYOLOv1.3-small at 150MHz would be the ideal choice or if the application needs the fastest inference time and most accuracy, running TinyissimoYOLOv8-Big at 370MHz is the ideal choice.

\begin{figure}[tb] 
\centering
\centering
\includegraphics[trim={0.1cm 0.1cm 0.1cm 0.1cm},clip, width=0.5\linewidth]{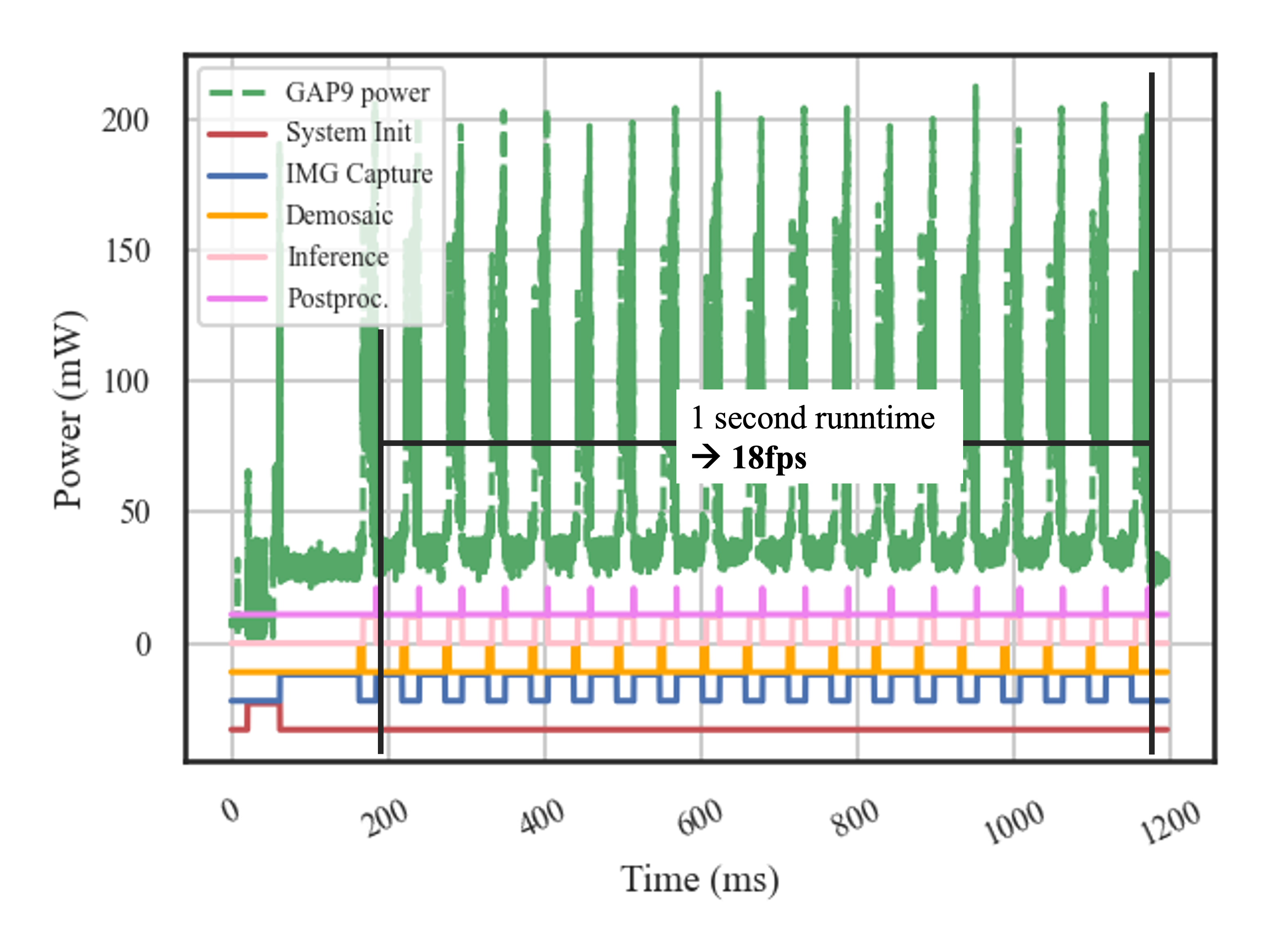}
\caption{\textbf{Full System Power Measurement:} The lines described in the legend show if the mentioned system process is running or not running. We show 18fps GAP9 on-device image-capturing, demosaicing, network inference and postprocessing execution.}
\label{fig:1second_demo_drawn} 
\end{figure}

\subsection{System Results}
Running the end-to-end solution on the GAP9 requires the system to initialize all required processing units of the \gls{soc}. Including the NE16 accelerator of GAP9, initializing the CSI-2 interface for the HM0360 camera from Himax, and creating all the needed memory buffers for image capturing and demosaicing/de-bayering. Once initialized, the system can run the system application loop, which consists of image capturing (double-buffered), demosaicing of the image, \gls{ai} inference execution (TinyissimoYOLOv1.3), and post-processing of the network output. \cref{fig:1second_demo_drawn} shows the power consumption during the full operation of the system, running the loop ten consecutive times. For the following power measurements, the loop was executed 100x and the average is reported. Capturing an image 100 times took on average \SI{34.69}{ms} while consuming \SI{1.17}{mJ} or \SI{18.79}{mA}. Demosaicing needed only \SI{4.87}{ms} while consuming \SI{23.82}{mA} resulting in \SI{0.209}{mJ} of energy. Running the small TinyissimoYOLOv1.3 network on NE16 consumed \SI{52.27}{mA} of current during \SI{16.86}{ms} resulting in \SI{1.59}{mJ} of Energy. The rest of the TinyissimoYOLO version is evaluated in \cref{sec:network_performance}. \cref{tab:demo_consumption} summarizes the current, power, and energy measurements for the corresponding demonstration process. Last, the post-processing took \SI{27}{\micro s}, consuming \SI{28.3}{mA} resulting in \SI{1.38}{\micro J}. With that, the average loop execution consumes \SI{3.28}{mJ}, i.e. \SI{29.55}{mA} for \SI{61.67}{ms}. 

\begin{table}[b!]
    \caption{\textbf{System demonstrator:} energy consumption measurement.}
    \centering
    \scriptsize
    \begin{tabular}{@{}llllll@{}} 
        & & \textbf{Current }& \textbf{Power} & \textbf{Energy} & \textbf{Time} \\
        &\textbf{Event} & \textbf{[mA]} & \textbf{[mW]} & \textbf{[mJ]} & \textbf{[ms]} \\
        \toprule
        &System quiescent$^*$ & 0.75 & 2.48 & 2.48 & \\
        & Init system & 11.96 & 21.54 & 0.89 & 41.44 \\
        \midrule
        \multirow{4}{*}{\rotatebox[origin=c]{90}{\textbf{Loop}}} &Capture image & 18.78 & 33.82 & 1.17 & 34.69 \\
        &Demosaic non-cluster & 23.82 & 42.88 & 0.21 & 4.87 \\
        &Run Tinyissimoyolov1.3 & 52.27 & 94.10 & 1.59 & 16.86 \\
        &Post processing & 28.26 & 50.86 & 0.001 & 0.03\\
        \midrule
        & \textbf{Loop AVG.}& 30.0 & 54.0 & 3.05 & 56.45 \\
        \bottomrule
        $^*$ & \multicolumn{5}{l}{nRF wakes up from sleep mode, changes power settings,} \\
        & \multicolumn{5}{l}{and sleeps again.} \\
    \end{tabular}
    \label{tab:demo_consumption}
\end{table}

\paragraph{Battery runtime estimation}
The smart glasses are charged with a battery, which fits inside the opposite temple of the smart-glasses' electronics, see \cref{fig:smart_glasses_rendering2}, with a maximal energy content of up to \SI{154}{mAh} and \SI{585}{Wh} with a nominal voltage level of around \SI{3.8}{V}. The continuous execution of the presented end-to-end system consumes \SI{30.0}{mA} at \SI{1.8}{V} continuously and draws \SI{54}{mW} from the battery. Including the HM0360 image sensor and the nRF \gls{mcu}, the total system power consumption is \SI{62.9}{mW}, resulting in a battery runtime of \SI{9.3}{h}.
In perspective the newly released \textit{RayBan-Meta} smart glasses, with the same sized battery capacity, claim to last 4 hours with moderate usage or up to 3 hours of continuous audio streaming and voice assistance\footnote{\url{https://www.meta.com/ch/en/legal/ray-ban-meta/disclosures/}}.

\section{Conclusion}
\label{sec:conclusion}

This paper proposed a novel smart glasses platform and demonstrated the system's capabilities. We perform image capturing and demosaicing, before running \gls{ai} inference and post-processing the networks' output to get bounding boxes. The end-to-end processing loop takes \SI{56}{ms} and consumes \SI{62.9}{mW} resulting in 18fps of continuous end-to-end execution for \SI{9.3}{h} on a battery with \SI{154}{mAh}. This sets a notable achievement for image processing on \gls{mcu} class devices.

Further, this paper proposed a family of new TinyissimoYOLO versions, using the YOLOv3 detection layer and the  \gls{yolo} version-specific head, while evaluating the architectures proposed in TinyissimoYOLOv1, YOLOv5, YOLOv8, and YOLOv10. The networks contain 50x to 100x fewer parameters than the initial YOLOv1 version and have been evaluated and compared on the PascalVOC and MS-COCO test datasets. The networks achieve sub-million parameters for up to 20 classes and fit quantized on \glspl{mcu}. TinyissimoYOLOv8 with \SI{840}{k} parameters, achieves 44\% \gls{map} while being executed within and \SI{34}{ms} on the GAP9. The fastest small TinyissimoYOLOv1.3 is executed within \SI{16.2}{ms} consuming only \SI{1.27}{mJ} of energy for one inference and achieves 30\% \gls{map}.
As such, this paper presents a highly generalized multi-class detection family of networks running with near SOTA performance detection accuracy in real-time ($>$18fps) on the GAP9 \gls{mcu}. 


\section*{Acknowledgment}
The authors would like to thank Dominik Müller for his commitment during his semester’s project. As well as armasuisse Science \& Technology, and the Swiss National Science Foundation (SNSF) project 200021E\_219943 Neuromorphic Attention Models for Event Data (NAMED) for funding this research.
\bibliographystyle{splncs04}
\bibliography{main}
\end{document}